\documentclass[runningheads]{llncs}

\usepackage{eccv}

\usepackage{eccvabbrv}

\usepackage{graphicx}
\usepackage{booktabs}
\usepackage{multirow}
\usepackage{changes}
\usepackage{marvosym}

\usepackage{colortbl}
\definecolor{mygray}{gray}{.9}

\usepackage[accsupp]{axessibility}  

\usepackage{hyperref}

\usepackage{orcidlink}

\begin{document}

\title{QueryCDR: Query-Based Controllable Distortion Rectification Network for Fisheye Images} 

\titlerunning{QueryCDR: Query-Based Controllable Distortion Rectification Network}

\author{Pengbo~Guo\inst{1,2}\orcidlink{0009-0003-2228-4550} \and
Chengxu~Liu\inst{2,3}\orcidlink{0000-0001-8023-9465} \and
Xingsong~Hou\inst{2}\textsuperscript{(\Letter)}\orcidlink{0000-0002-6082-0815} \and
Xueming~Qian\inst{2,3}\orcidlink{0000-0002-3173-6307}}

\authorrunning{P.~Guo et al.}

\institute{School of Software Engineering, Xi'an Jiaotong University, Xi'an, China \and Xi'an Jiaotong University, Xi'an, China \and 
Shaanxi Yulan Jiuzhou Intelligent Optoelectronic Tech. Co., Ltd, Xi'an, China\\
\email{\{guopengbo,chengxuliu\}@stu.xjtu.edu.cn, 
\{houxs,qianxm\}@mail.xjtu.edu.cn}}

\maketitle

\begin{abstract}

  Fisheye image rectification aims to correct distortions in images taken with fisheye cameras. Although current models show promising results on images with a similar degree of distortion as the training data, they will produce sub-optimal results when the degree of distortion changes and without retraining. The lack of generalization ability for dealing with varying degrees of distortion limits their practical application. In this paper, we take one step further to enable effective distortion rectification for images with varying degrees of distortion without retraining. We propose a novel Query-Based Controllable Distortion Rectification network for fisheye images (QueryCDR). In particular, we first present the Distortion-aware Learnable Query Mechanism (DLQM), which defines the latent spatial relationships for different distortion degrees as a series of learnable queries. Each query can be learned to obtain position-dependent rectification control conditions, providing control over the rectification process. Then, we propose two kinds of controllable modulating blocks to enable the control conditions to guide the modulation of the distortion features better. These core components cooperate with each other to effectively boost the generalization ability of the model at varying degrees of distortion. Extensive experiments on fisheye image datasets with different distortion degrees demonstrate our approach achieves high-quality and controllable distortion rectification. Code is available at \url{https://github.com/PbGuo/QueryCDR}.
  
\end{abstract}

\section{Introduction}

Benefiting from the huge field-of-view (FoV), fisheye cameras are widely utilized in various fields, including security surveillance~\cite{lin2012integrating,muhammad2018efficient} and autonomous driving~\cite{geiger2012we,rashed2021generalized}. However, the distortion brought by the fisheye lenses greatly limits the performance of downstream vision tasks~\cite{duan2020rapid,plaut20213d,kumar2021syndistnet,zhang2021automatic}. How to eliminate the distortion in fisheye images has attracted great attention in recent years.

\begin{figure}[tb]
    \centering
    \includegraphics[width=\textwidth]{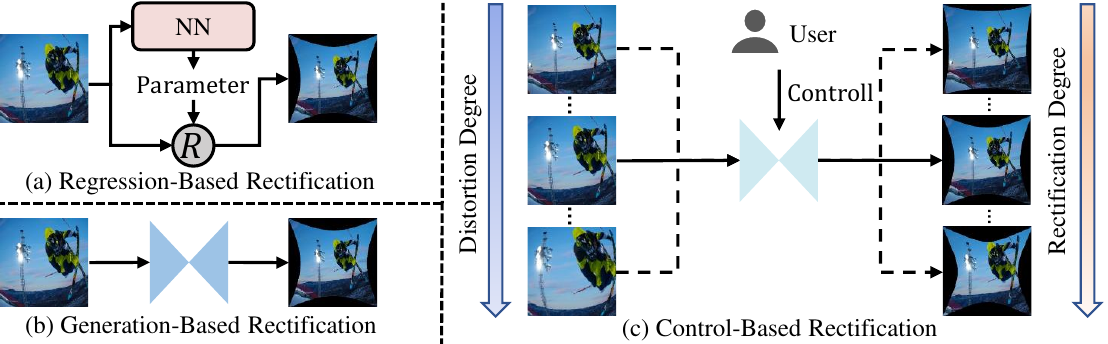}
    \caption{Different approaches to fisheye image distortion rectification. (a) Regression-Based: Using a neural network to predict distortion-related parameters, then apply rectification algorithms $R$ for rectification. (b) Generation-Based: Input the distorted fisheye image and directly generate the rectified image end-to-end. (c) Control-Based: Users provide control conditions to guide the rectification process, resulting in promising rectified images of various distortion degrees.}
    \label{fig:intro}
\end{figure}

Early methods~\cite{barreto2005fundamental,hartley2007parameter,henrique2013radial,puig2011calibration,bukhari2013automatic,wang2009simple,zhang2015line,santana2016iterative} primarily relied on identifying matching feature points or curves for automatic rectification. However, constrained by the instability of feature detection, both the generalization of the algorithm and the quality of the rectified images were unsatisfactory~\cite{liao2019dr,yang2021progressively,feng2023simfir}. In recent years, owing to the robust learning and generalization capabilities of neural networks, deep learning-based fisheye image rectification methods have become mainstream. These methods can be divided into two categories: 
regression-based rectification~\cite{rong2017radial,yin2018fisheyerecnet,bogdan2018deepcalib,li2019blind,xue2019learning} and generation-based rectification~\cite{liao2019dr,liao2020model,yang2021progressively,feng2023simfir,yang2023dual}. The former uses deep regression models to predict distortion parameters for image reconstruction, as shown in~\cref{fig:intro}(a). The latter uses an encoder-decoder structure to generate well-rectified images directly, as shown in~\cref{fig:intro}(b).

It is worth noting that these methods only achieve satisfactory results on a similar degree of distortion as the training data. This means that when handling different degrees of distortion without retraining, there will be a significant decrease in the quality of the rectified images~\cite{liao2020model}. This is because the model tends to learn fixed position mapping relationships during training. When the distortion degree changes, this relationship does not work for the new distortion distribution. Therefore, these models usually need to be retrained on images with different distortion degrees. In addition, fisheye image acquisition is difficult, and re-collecting new datasets with varying degrees of distortion would incur significant costs~\cite{yang2021progressively,yang2023dual}. Thus, it is essential to explore a model that can handle different degrees of distortion simultaneously.

In recent years, some methods have been proposed to achieve controllable image restoration~\cite{wang2019deep,he2019modulating,wang2019cfsnet,he2020interactive,cai2021toward,mou2022metric,zhang2023real,yao2023towards}. Typically, CFSNet~\cite{wang2019cfsnet}, MM-RealSR~\cite{mou2022metric}, and Yao \etal~\cite{yao2023towards} introduce scalars as control conditions to effectively restore degradation of varying degrees.
It would be a promising solution to introduce an effectively controllable mechanism to deal with all degrees of distortion, as shown in~\cref{fig:intro}(c). However, directly applying existing controllable mechanisms in restoration tasks to the fisheye rectification model suffers from the following challenges: 
1) There is a gap between optimization objectives (\ie~the roles of the controllable mechanisms). Restoration models learn the pixel-level detail restoration, whereas fisheye rectification models learn the spatial-level positional mapping relationships~\cite{liao2020model,feng2023simfir}. The control mechanism lacking position information in restoration tasks is ineffective in controlling distortion rectification networks. 
2) There is a gap between optimization difficulties (\ie~the roles of the control conditions). The distortion of a fisheye image increases gradually from the image center to boundary~\cite{yang2021efficient}. Therefore, it is not suitable to use a single scalar control condition in restoration tasks to deal with such spatially varying distortions. These challenges restrict the application of controllable mechanisms and control conditions in fisheye image rectification.

To address these issues, we propose the Query-Based Controllable Distortion Rectification network (QueryCDR), as shown in Fig.~\ref{fig:overview}. By introducing a series of learnable queries as control conditions, QueryCDR allows the users to achieve fisheye rectification with different distortion degrees. Specifically, to incorporate positional mapping relationships into control conditions, we introduce the Distortion-aware Learnable Query Mechanism (DLQM), which defines a series of queries representing different rectification control conditions. During inference, DLQM extracts position-dependent control conditions from the user-given query and feeds them into the network for controlling the rectification process. Furthermore, to enable the control conditions to guide the rectification efficiently, we propose two types of controllable modulating blocks: the Controllable Convolution Modulating Block (CCMB) based on CNN~\cite{he2016deep}, and the Controllable Attention Modulating Block (CAMB) based on Transformer~\cite{vaswani2017attention}. They are good at extracting local texture features and learning long-range distortion mapping relationships, respectively. By combining CCMB and CAMB, we construct a robust controllable rectification network. Our QueryCDR can handle various distortions without retraining, enhancing the generalization ability of the fisheye rectification model.

We summarize our contributions as follows:
\begin{itemize}
    \item We propose QueryCDR, a Query-Based Controllable Distortion Rectification network for fisheye images. Extensive experiments demonstrate that our QueryCDR can deliver superior results on a variety of distortion degrees.
    \item We propose the Distortion-aware Learnable Query Mechanism (DLQM), which effectively introduces the latent spatial relationships to control conditions for fisheye image rectification.
    \item We propose two kinds of blocks for modulating features using control conditions: the Controllable Convolution Modulating Block (CCMB) and the Controllable Attention Modulating Block (CAMB). They can effectively utilize control conditions to guide the rectification process.
\end{itemize}

\section{Related Work}
\subsection{Traditional Fisheye Image Rectification}

Traditional rectification methods can be divided into two types, multi-view-based and line-based methods. Multi-view-based methods~\cite{barreto2005fundamental,hartley2007parameter,henrique2013radial,kukelova2011minimal,puig2011calibration,scaramuzza2006flexible,sturm2004generic} calibrated fisheye images by finding corresponding feature points from multiple viewpoints. Line-based methods~\cite{mei2007single,devernay2001straight,barreto2005geometric,bukhari2013automatic,thormahlen2003robust,wang2009simple,zhang2015line,santana2016iterative} employed line detection to rectify the curved lines, thereby achieving distortion rectification. However, these methods require manual intervention or handcrafted feature extractors, and the rectification process is unstable, failing to achieve satisfactory results.

\subsection{Deep Learning Based Fisheye Image Rectification}

Rapid advances in deep learning have allowed them to shine in low-level vision tasks~\cite{cai2016dehazenet,ledig2017photo,zhang2017beyond,wang2018esrgan,liu2022learning,liu2023fsi,liu20234d,liu2024motion}. As a core task in low-level vision, distortion image rectification has received increasing attention~\cite{rong2017radial,yin2018fisheyerecnet,liao2019dr,yang2021progressively,feng2023simfir,yang2023dual}. According to the network architecture, deep learning-based distortion image rectification methods can be categorized into two types, regression-based methods~\cite{rong2017radial,yin2018fisheyerecnet,bogdan2018deepcalib,li2019blind,xue2019learning} and generation-based methods~\cite{liao2019dr,liao2020model,yang2021progressively,feng2023simfir,yang2023dual}. 

Regression-based methods~\cite{rong2017radial,yin2018fisheyerecnet,bogdan2018deepcalib,li2019blind,xue2019learning} utilize neural networks to predict distortion-related coefficients for rectifying the image. Rong \etal~\cite{rong2017radial} were the first to use CNNs for fisheye image rectification. They employed the network to predict across multiple distortion intervals, achieving preliminary rectification results. Yin \etal~\cite{yin2018fisheyerecnet} integrated semantics as prior information to guide rectification. While these methods have shown some performance improvement, they face the challenge of non-end-to-end design, thus requiring a large number of additional labels, and increasing operational costs. Therefore, to reduce models' complexity, Generation-based methods~\cite{liao2019dr,liao2020model,yang2021progressively,feng2023simfir,yang2023dual} employ generative networks to take distorted images as input and directly generate rectified images. DR-GAN~\cite{liao2019dr} was the first to apply a generative adversarial network (GAN)~\cite{goodfellow2020generative} to fisheye image rectification, enabling the network to directly generate rectified images without estimating additional parameters. DDM~\cite{liao2020model} introduced distortion maps to help the model better learn the distortion distribution. PCN~\cite{yang2021progressively} designed a flow estimation module to predict appearance flows in fisheye images, using it to assist the rectification progress. SimFIR~\cite{feng2023simfir} introduced a self-supervised rectification module, allowing the network to better learn the distortion representations at different locations. However, both paradigms require retraining when handling images with different distortion degrees, failing to address the issue of weak model generalization.

\subsection{Controllable Low-level Vision}

To address the issue of varying degradation levels in low-level vision tasks, an increasing number of studies~\cite{wang2019deep,he2019modulating,wang2019cfsnet,he2020interactive,cai2021toward,mou2022metric,zhang2023real,yao2023towards} propose controllable network architectures to tackle this challenge. DNI~\cite{wang2019deep} interpolated all parameters of different restoration networks, which were trained with different degradation levels. By adjusting the interpolation coefficients, a smooth control of the image can be achieved. AdaFM~\cite{he2019modulating} achieved better results by inserting AdaFM layers after each convolution layer to change the filters' statistics, thus the users can interactively manipulate the restoration results by tuning a control coefficient. CFSNet~\cite{wang2019cfsnet} introduced the tuning branch to adaptively learn the control coefficients, and then use them to couple the features with the main branch. MM-RealSR~\cite{mou2022metric} proposed a metric learning strategy to map unquantifiable degradation levels to a metric space as control conditions. However, due to the different optimization objectives and complexity of the control mechanisms, directly applying these methods to fisheye image distortion rectification tasks will not achieve effective control over the rectification process.

\section{Methodology}

\begin{figure}[tb]
    \centering
    \includegraphics[width=\linewidth]{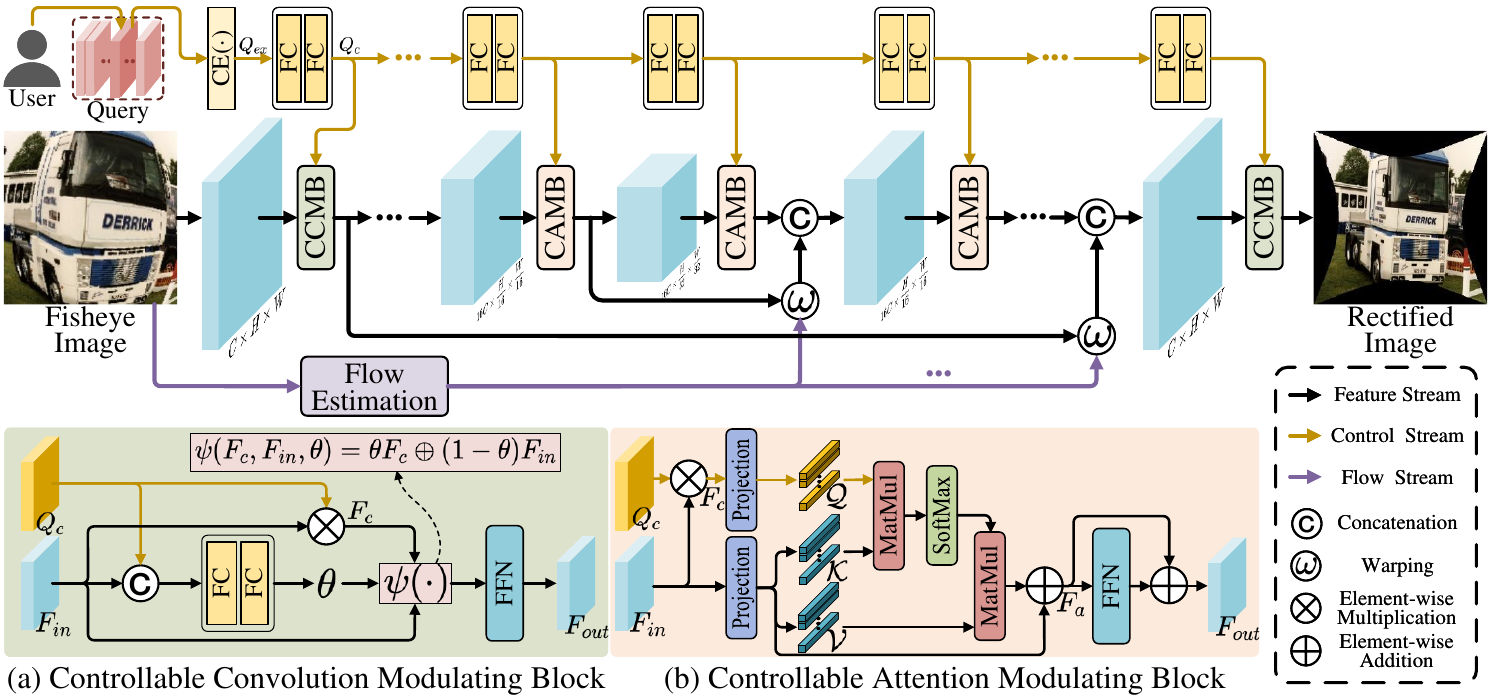}
    \caption{Overview of our proposed Query-Based Controllable Distortion Rectification network (QueryCDR). The Distortion-aware Learnable Query Mechanism (DLQM) extracts control conditions from user-given queries and feeds them layer by layer into the rectification network. The rectification network is composed of Controllable Convolution Modulating Blocks (CCMB) and Controllable Attention Modulating Blocks (CAMB), which modulate the input features $F_{in}$ with control conditions $F_c$, enabling controllable rectification process.}
    \label{fig:overview}
\end{figure}

\subsection{Overview}

The overview of our proposed Query-Based Controllable Distortion Rectification network (QueryCDR) is shown in~\cref{fig:overview}. First, following existing work~\cite{yang2021progressively}, the input fisheye image is fed into the flow estimation module to obtain the appearance flow, which performs a coarse-grained rectification for image features (\ie, warping $\omega(\cdot)$). Then, the Distortion-aware Learnable Query Mechanism (DLQM) (in~\cref{sec:vdm}) extracts control conditions from the user-given query and feeds them layer by layer into the rectification network. Finally, a U-shaped hierarchical network composed of several controllable modulating blocks (in~\cref{CMB}) is used to rectify the distorted input image. These blocks modulate features with the control conditions given by DLQM, to get the final output.

In the following, we will provide a comprehensive description of each module and its corresponding role in the QueryCDR.

\subsection{Distortion-aware Learnable Query Mechanism (DLQM)}
\label{sec:vdm}

\begin{figure}[tb]
    \centering
    \includegraphics[width=\textwidth]{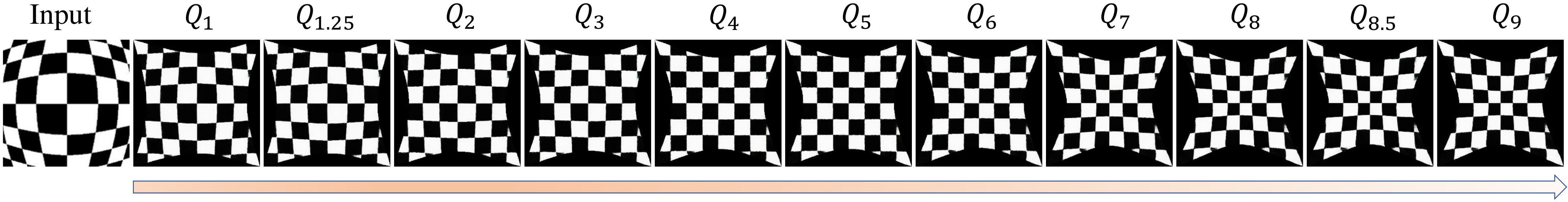}
    \caption{Given an input image, QueryCDR can accurately produce results with different rectification degrees by feeding different queries. Moreover, by interpolating between different queries, we can achieve smooth continuous rectification for any distortion degree. For examples, $Q_{1.25}=0.75Q_1+0.25Q_2$, and $Q_{8.5}=0.5Q_8+0.5Q_9$.}
    \label{fig:chessboard}
\end{figure}

To achieve control over the rectification process, most existing controllable methods~\cite{wang2019cfsnet,mou2022metric,yao2023towards} introduce scalars as control conditions to represent the degradation degrees for image reconstruction. However, due to the peculiarities of distortion in fisheye images~\cite{yang2022fishformer,shen2022panoformer,feng2023simfir}, these methods fail to achieve effective control over the rectification process. To address this issue, we propose the DLQM. It maintains a learnable query set, providing diverse effective control conditions for the rectification network. During training, DLQM projects the learned high-dimensional positional mapping relationship into a low-dimensional latent space of queries. During inference, DLQM extracts the control information from the queries and converts it into corresponding control conditions for each layer of the rectification network. Simply by providing different queries, DLQM can effectively control the rectification process, as shown in~\cref{fig:chessboard}. At the same time, our QueryCDR can also achieve smooth continuous rectification for any distortion degrees by simply interpolating between different queries.

Specifically, we construct a query set $\mathbf{Q}_s = \{Q_i\mid Q_i\in \mathbb{R}^{C_{in}\times H_{in}\times W_{in}}, i = 1, 2, \ldots, N\}$, which comprises $N$ queries representing different distortion degrees. $H_{in}$, $W_{in}$, and $C_{in}$ represent the query's height, width, and channel, respectively. Each query has the same size as the input image $I_{in}\in \mathbb{R}^{C_{in}\times H_{in}\times W_{in}}$. During inference, users select a query $Q_i$ from $\mathbf{Q}_s$ and feed it into the DLQM. In DLQM, the control extracting part $\text{CE}(\cdot)$ is firstly used to extract the features in $Q_i$. The input processing part can be expressed as follows,
\begin{equation}Q_{ex} = \text{CE}(Q_i),\end{equation}
where $Q_{ex}\in\mathbb{R}^{C_{in}\times H_{in}\times W_{in}}$ is the extracted feature, and serves as the input to the first control layer of DLQM. To minimize the computational costs and effectively extract the control information, $\text{CE}(\cdot)$ is composed of three convolution layers with a kernel size of $3 \times 3$.

Subsequently, DLQM provides corresponding control conditions to the respective layers of the rectification network. To provide appropriate control conditions, each control layer of DLQM comprises two fully connected layers $\text{FC}_1(\cdot)$, $\text{FC}_2(\cdot)$. The $l$-th layer of DLQM can be represented as follows,
\begin{equation}Q_c^{l} = \text{FC}_2^l(\text{FC}_1^l(Q_c^{l-1})),\end{equation}
where $Q_c^{l-1}$ is the output control condition from the $(l-1)$-th control layer. For $l=1$, $Q_c^0$ represents the extracted feature $Q_{ex}$. The output $Q_c^{l}$ of the $l$-th control layer is fed into the $l$-th rectification layer as a control condition, and simultaneously serves as input to the $(l+1)$-th control layer of DLQM.

Afterward, we feed the control conditions into the rectification network. Each layer's controllable modulating block modulates the input feature with the control condition, guiding the modulated features toward the desired direction.

\subsection{Controllable Modulating Block}
\label{CMB}

When controlling the rectification process, to avoid obtaining results with blurred texture details or residual distortions, we introduce two types of controllable modulating blocks tailored for reconstructing local texture details and learning continuous distortion patterns. They are denoted as the Controllable Convolution Modulating Block (CCMB) based on CNN~\cite{he2016deep} (in~\cref{fig:overview}(a)) and the Controllable Attention Modulating Block (CAMB) based on Transformer~\cite{vaswani2017attention} (in~\cref{fig:overview}(b)). The CCMB can adaptively fuse the original features and the controlled features, preserving more local details. The CAMB can better capture the spatial distortion information, guaranteeing the integrity of the recovered content~\cite{peng2021conformer,yang2022fishformer,feng2023simfir}. To balance performance and computational cost, our QueryCDR constructs a U-shaped rectification network~\cite{yang2021progressively} composed of CCMB and CAMB. In the first three layers with larger feature maps, \ie, $l=\{1,2,3,9,10,11\}$, CCMB is employed to learn more local texture details. In the remaining layers with smaller feature maps, \ie, $l=\{4,5,6,7,8\}$, CAMB is utilized to capture more global dependencies within the images.

\subsubsection{Controllable Convolution Modulating Block (CCMB).}

When incorporating control conditions into the rectification network, existing methods~\cite{wang2019cfsnet,yao2023towards} fail to effectively balance the fusion ratio between original features and controlled features. Directly using the controlled features or fusing them with original features at a fixed ratio will degrade the quality of the rectified images. To harmoniously incorporate control conditions into the rectification process, we design the CCMB, as shown in~\cref{fig:overview}(a). CCMB can dynamically find an optimal ratio to modulate the distorted features with control conditions.

Specifically, CCMB receives an input feature $F_{in}\in \mathbb{R}^{C \times H \times W}$ and a control condition $Q_c\in \mathbb{R}^{C \times H \times W}$. Both of them are used to predict the fusion ratio. Here we omit the layer information for brevity,
\begin{equation}\theta = \text{CP}(F_{in}, Q_c),\end{equation}
where $\text{CP}(\cdot)$ is the coefficient predictor composed of two fully connected layers, takes the concatenation of $F_{in}$ and $Q_c$ as input and predicts the fusion ratio $\theta$.

Then, we perform element-wise multiplication $\otimes$ on $F_{in}$ and $Q_c$ to yield the controlled features $F_c\in \mathbb{R}^{C \times H \times W}$, which can be expressed as,
\begin{equation}\label{equ:controlled_fea} F_c = F_{in} \otimes Q_c.\end{equation}
Finally, the input features $F_{in}$ and controlled features $F_c$ are combined in a weighted sum according to the fusion ratio $\theta$, illustrated as $\psi(\cdot)$ in~\cref{fig:overview}(a),
\begin{equation}F_{out} = \psi(F_c,F_{in},\theta) = \theta F_c \oplus (1- \theta)F_{in}, \end{equation}
where $\oplus$ represents the element-wise addition, and $F_{out}\in \mathbb{R}^{C \times H \times W }$ is the final output of CCMB.

Due to this dynamic modulation mechanism, CCMB achieves effective control over the rectification process while preserving richer texture details.

\subsubsection{Controllable Attention Modulating Block (CAMB).}

Due to the lack of perception of global information, CNN-based networks struggle to learn long-range dependencies, particularly the continuous and amorphous distortions prevalent in fisheye distortions~\cite{yang2022fishformer,shen2022panoformer,feng2023simfir}. In contrast, Transformer-based networks effectively compensate for this, with their global attention mechanism~\cite{peng2021conformer,guo2022cmt}.

Therefore, we propose the CAMB, as illustrated in~\cref{fig:overview}(b). To optimally leverage the control conditions given by DLQM, we designed the control-attention mechanism, enabling CAMB to perceive the global spatial relationships in the control conditions effectively. 
Specifically, we unfold and project the controlled feature $F_c$ (in~\cref{equ:controlled_fea}) as the query $\mathcal{Q}\in \mathbb{R}^{m\times L}$, with the input feature $F_{in}$ as the key $\mathcal{K}\in \mathbb{R}^{m\times L}$ and value $\mathcal{V}\in \mathbb{R}^{m\times L}$, where $L = H\times W$ represents the sequence length, and $m$ denotes the dimensions of the sequences. The control-attention is described by the following,
\begin{equation} 
\begin{split}
    & \mathcal{Q}=W_\mathcal{Q}F_c, \mathcal{K}=W_\mathcal{K}F_{in}, \mathcal{V}=W_\mathcal{V}F_{in},\\
    & \text{CTRL-ATTN}(\mathcal{Q},\mathcal{K},\mathcal{V})=\text{softmax}(\frac{\mathcal{Q}\mathcal{K}^T}{\sqrt{m}})\mathcal{V},
\end{split}
\end{equation}
where $W_\mathcal{Q}, W_\mathcal{K}, W_\mathcal{V} \in \mathbb{R}^{m\times C}$ represent the projection matrices of the queries, keys, and values, respectively. $\text{CTRL-ATTN}(\cdot)$ represents the control-attention we proposed. The overall computation of CAMB can be formulated as follows,
\begin{equation}
\begin{split}
    & F_a =\text{CTRL-ATTN}(\text{LN}(\mathcal{Q},\mathcal{K},\mathcal{V}))\oplus F_{in},\\
    & F_{out} =\text{Conv}_{1 \times 1}(\text{FFN}(\text{LN}(F_a))\oplus F_a),
\end{split}
\end{equation}
where $F_{in}$ means the input feature, $F_a\in \mathbb{R}^{m\times H\times W}$ and $F_{out}\in \mathbb{R}^{C\times H\times W}$ are the outputs of $\text{CTRL-ATTN}(\cdot)$ and CAMB, respectively. $\text{LN}(\cdot)$ denotes the layer normalization~\cite{ba2016layer}. $\text{FFN}(\cdot)$ stands for the feed-forward network composed of three fully connected layers, which helps CAMB to focus on the global dependencies.

CAMB can discern the global mapping relationships within fisheye images, ensuring rectification uniformity compared to the CNN-based networks. 

\subsection{Training Strategy}
\label{sec.train}

To guarantee the robust and stable controllable distortion rectification of Query-CDR, we design a two-stage training strategy that combines coarse-grained distortion pre-training and fine-grained distortion fine-tuning.

During the coarse-grained distortion pre-training phase, we choose the most commonly used dataset~\cite{liao2019dr,yang2021progressively,feng2023simfir,yang2023dual} that only contains one degree of distortion to train our QueryCDR. For clarity, we denote the distortion degree in this phase as $d$. Correspondingly, only one query, denoted as $Q$, is used for training. The optimization objective can be expressed as follows,
\begin{equation}
\mathcal{L}_{pre}=\mathcal{L}_r+\mathcal{L}_{m},
\end{equation}
where $\mathcal{L}_{pre}$ is the overall loss function for the pre-training phase. $\mathcal{L}_r$ denotes the reconstruction loss,
\begin{equation}
\mathcal{L}_r=\left\|I_{out}^{d}-I_{gt}^{d}\right\|_1,\\
\end{equation}
where $I_{out}^{d}$ and $I_{gt}^{d}$ signifie the output result and ground truth, respectively. $\mathcal{L}_{m}$ denotes the multi-scale loss,
\begin{equation}
\mathcal{L}_m=\sum_{j=1}^{Z-1}\|S(I_{gt}^{d},j)-C(F_{out}^j)\|_1,
\end{equation}
where $S(\cdot)$ represents the operation that down-samples the input $I_{gt}^{d}$ by a factor of $1/2^j$. $Z$ represents the number of decoder's layers, and we set $Z$ to 6. $F_{out}^j$ denotes the feature in $j$-th decoder layer. $C(\cdot)$ is $3\times3$ convolution for decoding the features into 3-channel RGB images. In this way, each feature map on the decoder can be effectively supervised. 
The $I_{out}^{d}$ can be obtained as,
\begin{equation}
I_{out}^{d}=\text{QueryCDR}(I_{in}^{d},Q),
\end{equation}
where $I_{in}^{d}$ represents fisheye images with distortion degree $d$ as input. This way can effectively boost the model stability and accelerate the convergence.

During the fine-grained distortion fine-tuning phase, we use the varying distortion degrees datasets to fine-tune our QueryCDR. Before training, We replicate the weight of the query $Q$ to the other queries to accelerate convergence on other distortion degrees. Subsequently, we fine-tune our QueryCDR using fisheye images with varying distortion degrees, and feeding corresponding queries into QueryCDR for training at the same time. This allows the query set to efficiently acquire diverse latent spatial relationships. The optimization objective can be expressed as follows,
\begin{equation}
\mathcal{L}_{fine}^{d_i}=\mathcal{L}_r^{d_i}+\mathcal{L}_{m}^{d_i},
\end{equation}
where $\mathcal{L}_{fine}^{d_i}$ is the overall fine-tuning loss function for distortion degree $d_i, i\in\{1,2,\cdots,9\}$, which is similar to the $\mathcal{L}_{pre}$ but calculated across different distortion degrees. The $I_{out}^{d_i}$ can be obtained as,
\begin{equation}
    I_{out}^{d_i}=\text{QueryCDR}(I_{in}^{d_i},Q_{i}),
\end{equation}
where $I_{in}^{d_i}$ denotes fisheye images with a distortion degree of $d_i$ as input, $Q_{i}$ denotes the query corresponding to $d_i, i\in\{1,2,\cdots,9\}$.

With this two-stage training strategy, our QueryCDR can effectively utilize only a small amount of varying distortion degrees data that is difficult to obtain, to rapidly finish the training of our network.

\begin{table*}[tb]\small
    \centering
    \caption{Quantitative comparison (PSNR (dB)$\uparrow$, SSIM $\uparrow$) on COCO~\cite{lin2014microsoft} fisheye image dataset with varying distortion degrees. \textcolor{red}{Red} indicates the best and \textcolor{blue}{blue} indicates the second best performance (best viewed in color).}
    \label{Performance_table_psnr}
    \begin{tabular}{l*{10}{c}}
    \toprule
    \multirow{2}{*}{Method} & \multicolumn{10}{c}{PSNR}  \\
    \cmidrule(lr){2-11}
    \rowcolor{mygray}
    &$d_1$ & $d_2$ & $d_3$ & $d_4$ & $d_5$ & $d_6$ & $d_7$ & $d_8$ & $d_9$ & Avg\\
    \midrule
    SC\cite{santana2016iterative} & 10.05 & 11.59 & 11.81 & 11.03 & 11.50 & 10.19 & 11.11 & 10.19 & 9.14 & 10.73 \\
    \rowcolor{mygray}
    DeepCalib\cite{bogdan2018deepcalib} & 10.00 & 10.69 & 11.01 & 11.19 & 11.28 & 11.46 & 11.45 & 11.13 & 11.11 & 11.04   \\
    Blind\cite{li2019blind}& 12.98 & 11.24 & 10.30 & 9.62 & 9.99 & 11.75 & 12.69 & 12.75 & 12.80 & 11.57   \\
    \rowcolor{mygray}
    DR-GAN\cite{liao2019dr} & 15.68 & 17.40 & 17.97 & 18.34 & 18.50 & 18.44 & 17.95 & 17.94 & 17.47 & 17.74   \\
    PCN\cite{yang2021progressively}& 14.93 & 17.43 & 18.43 & 18.86 & 18.86 & 18.88 & 18.74 & 17.35 & 18.26 & 17.97\\
    \rowcolor{mygray}
    DDA\cite{yang2023dual}& 16.39 & 17.41 &  17.43 & \textcolor{blue}{19.48} & \textcolor{blue}{20.12} & 18.90 & 18.85 & 18.17 & 18.22 & 18.33\\
    SimFIR\cite{feng2023simfir} & \textcolor{blue}{16.57} & \textcolor{blue}{17.88} & \textcolor{blue}{18.43} & 18.97 & 19.31 & \textcolor{blue}{19.28} & \textcolor{blue}{19.19} & \textcolor{blue}{18.65} & \textcolor{blue}{18.48} & \textcolor{blue}{18.53}\\
    \midrule
    \rowcolor{mygray}
    \textbf{QueryCDR} & \textcolor{red}{20.01}& \textcolor{red}{20.29}& \textcolor{red}{20.39} & \textcolor{red}{20.41} & \textcolor{red}{20.72} & \textcolor{red}{20.81} & \textcolor{red}{20.58} & \textcolor{red}{19.11} & \textcolor{red}{20.53} & \textcolor{red}{20.32}\\
    \midrule
    \multirow{2}{*}{Method} & \multicolumn{10}{c}{SSIM}  \\
    \cmidrule(lr){2-11}
    \rowcolor{mygray}
    &$d_1$ & $d_2$ & $d_3$ & $d_4$ & $d_5$ & $d_6$ & $d_7$ & $d_8$ & $d_9$ & Avg\\
    \midrule
    SC\cite{santana2016iterative} & 0.101 & 0.113 & 0.149 & 0.182 & 0.283 & 0.175 & 0.141 & 0.126 & 0.093 & 0.151 \\
    \rowcolor{mygray}
    DeepCalib\cite{bogdan2018deepcalib} & 0.184 & 0.210 & 0.223 & 0.230 & 0.234 & 0.246 & 0.250 & 0.245 & 0.246 & 0.229   \\
    Blind\cite{li2019blind}& 0.308 & 0.244 & 0.199 & 0.176 & 0.194 & 0.296 & 0.367 & 0.395 & 0.420 & 0.289   \\
    \rowcolor{mygray}
    DR-GAN\cite{liao2019dr} & 0.295 & 0.330 & 0.339 & 0.344 & 0.344 & 0.332 & 0.314 & 0.312 & 0.299 & 0.323   \\
    PCN\cite{yang2021progressively}& 0.420 & 0.547 & 0.589 & 0.607 & 0.608 & 0.610 & 0.615 & 0.576 & 0.603 & 0.575\\
    \rowcolor{mygray}
    DDA\cite{yang2023dual}& 0.455 & \textcolor{blue}{0.589} & 0.592 & 0.620 & \textcolor{blue}{0.675} & 0.626 & 0.619 & 0.564 & 0.581 & 0.591\\
    SimFIR\cite{feng2023simfir}&  \textcolor{blue}{0.492} & 0.581 & \textcolor{blue}{0.626} & \textcolor{blue}{0.635} & 0.640 & 0.628 & \textcolor{blue}{0.622} & \textcolor{blue}{0.591} & \textcolor{blue}{0.595} & \textcolor{blue}{0.601}\\
    \midrule
    \rowcolor{mygray}
    \textbf{QueryCDR} & \textcolor{red}{0.643} & \textcolor{red}{0.665} & \textcolor{red}{0.668} & \textcolor{red}{0.677} & \textcolor{red}{0.688} & \textcolor{red}{0.699} & \textcolor{red}{0.692} & \textcolor{red}{0.656} & \textcolor{red}{0.693} & \textcolor{red}{0.676}\\
    \bottomrule
    \end{tabular}
\end{table*}

\section{Experiment}
\label{exp}

\subsection{Experimental Settings}
\label{expset}

To demonstrate the effectiveness of our proposed QueryCDR, we followed the existing works~\cite{liao2019dr,yang2021progressively,feng2023simfir,yang2023dual}, employed the four-parameter polynomial model to synthesize fisheye images. We constructed synthetic datasets based on the original images of COCO~\cite{lin2014microsoft} and Places2~\cite{zhou2017places} datasets, respectively. Specifically, for images with distortion degree $d$, 40,000 images were used for the pre-training stage. And for images with varying distortion degrees $d_i, i\in\{1,2,\cdots,9\}$, 18,000 images were used for the fine-tuning stage, and 3,600 images for testing. The images were resized to $256\times256$ when fed into the network. For a fair comparison, we followed existing works~\cite{liao2019dr,yang2021progressively,feng2023simfir,yang2023dual} utilizing a batch size of 16 and the Adam~\cite{kingma2014adam} optimizer with a learning rate of 1e-4.

\begin{figure}[tb]
    \centering
    \includegraphics[width=\textwidth]{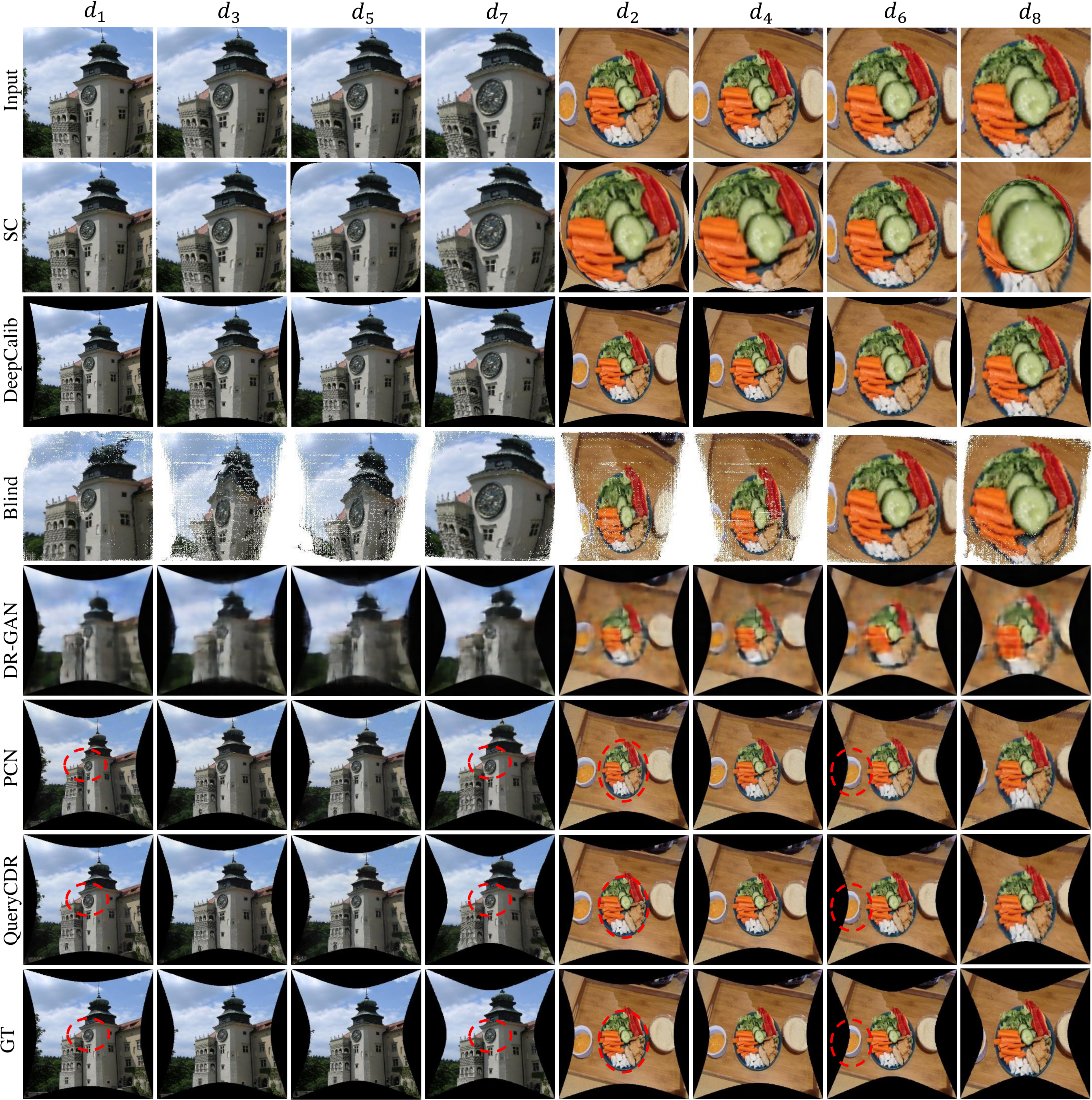}
    \caption{Qualitative results on synthetic fisheye images.}
    \label{fig:results}
\end{figure}

\subsection{Performance Evaluation}

To evaluate the performance of our method, we retrained and validated existing fisheye image rectification methods on synthetic fisheye datasets with 9 distortion degrees. The methods can be summarized into three categories: traditional method SC~\cite{santana2016iterative}, regression-based methods DeepCalib~\cite{bogdan2018deepcalib} and Blind~\cite{li2019blind}, and generation-based DR-GAN~\cite{liao2019dr}, PCN~\cite{yang2021progressively}, DDA~\cite{yang2023dual} and SimFIR~\cite{feng2023simfir}. For a fair comparison, we followed existing works~\cite{liao2019dr,yang2021progressively,yang2023dual,feng2023simfir} employing Peak Signal-to-Noise Ratio (PSNR) and Structural Similarity (SSIM) to quantify performance.

\subsubsection{Quantitative Results.}

Our performance comparisons are shown in~\cref{Performance_table_psnr}. Compared to existing methods, QueryCDR achieves the best performance across all distortion degrees without retraining. It is because DLQM effectively utilizes the control information in queries to guide the network in achieving varying degrees of rectification. Additionally, the capability of CCMB and CAMB allows for controlled rectification while obtaining high-quality results. Moreover, we observe that even on images with distortion degree $d_5$, which the existing methods perform well~\cite{liao2019dr,yang2021progressively,yang2023dual,feng2023simfir}, QueryCDR still outperforms the second best method~\cite{yang2023dual} by \textbf{0.60} dB of PSNR and \textbf{0.013} of SSIM. This further demonstrates the outstanding generalization ability of QueryCDR.

\subsubsection{Qualitative Results.}

\begin{figure}[tb]
    \centering
    \includegraphics[width=\textwidth]{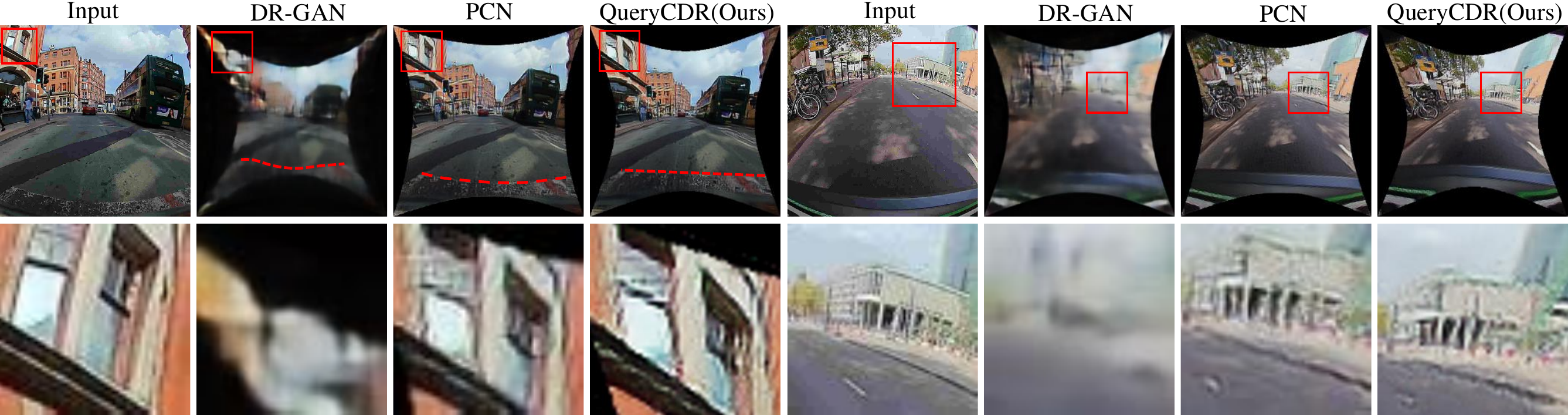}
    \caption{Qualitative results on real-world fisheye images.}
    \label{fig:results_rw}
\end{figure}

To further compare the visual qualities of different methods, we show the results rectified by QueryCDR and other rectification methods in~\cref{fig:results}. It can be observed intuitively that other methods fail to effectively rectify images with varying degrees of distortion. Particularly the objects in images, our QueryCDR maintains the accurate structure of objects across different distortions. Furthermore, benefiting from the modulation capability of CCMB and CAMB, QueryCDR preserves richer texture details after rectification.

In addition, to further demonstrate the generalization ability of QueryCDR, we conducted experiments on real-world fisheye image datasets~\cite{yogamani2019woodscape} in~\cref{fig:results_rw}. Despite the disparities between synthetic and real-world datasets, our QueryCDR still shows robust rectification capabilities, and preservers richer texture details. These results further validate its practicality in real-world scenarios.

\subsection{Ablation Study}
In this section, we conduct ablation for different control conditions, and study the effect of the controllable modulating blocks and the architecture setting.

\subsubsection{Distortion-aware Learnable Query Mechanism (DLQM).}
\label{sec.vector}

To validate the effectiveness of DLQM in controlling the rectification process, we implemented different rectification networks controlled by scalar, fixed query, and our learnable query in~\cref{abl_scalar}. When using scalar to control, the value of the scalar increments with the distortion of the entire image. When using fixed query to control, the parameters of the query are set to incremental values from 0 to 1 with the degree of distortion from the center to the edges. Following the settings in~\cref{expset}, we use the learnable query $Q_{i}, i\in \{1,2,\cdots,9\}$ to learn the distortion distribution corresponding to $d_i, i\in\{1,2,\cdots,9\}$.

We trained different methods using the same strategy as illustrated in~\cref{sec.train}, the results are shown in~\cref{abl_scalar} and~\cref{fig:abl_dlqm}. All three controllable methods outperform the uncontrollable method significantly at various degrees of distortion. This is because the controllable mechanism effectively assists the rectification network in distinguishing between different degrees of distortion, thereby enhancing the generalization capability of the rectification network. The improvement brought by scalar demonstrates the crucial role of controllable mechanisms in handling fisheye images with different degrees of distortion. The further enhancement with fixed query validates the importance of employing higher-dimensional control conditions for fisheye image rectification. Lastly, our tailored learnable query significantly outperforms all other controllable methods, verifying the unique superiority of our approach in controllable rectification.

\subsubsection{Controllable Modulating Block.}
\label{sec.abl_ccmb}

\begin{table*}[tb]
\begin{center}
    \caption{Performance comparison of different control methods.}
        \label{abl_scalar}
        \resizebox{\textwidth}{!}{
        \begin{tabular}{l c c c c c c c c c c}
            \toprule
            \multirow{2}{*}{Method} & \multicolumn{10}{c}{PSNR}  \\
            \cmidrule(lr){2-11}
            \rowcolor{mygray}
            &$d_1$ & $d_2$ & $d_3$ & $d_4$ & $d_5$ & $d_6$ & $d_7$ & $d_8$ & $d_9$ & Avg\\
            \midrule
            W/o Control & 14.93 & 17.43 & 18.43 & 18.86 & 18.86 & 18.88 & 18.74 & 17.35 & 18.26 & 17.97\\
            \rowcolor{mygray}
            Scalar & 19.52 & 20.25 & 19.97 & 19.43 & 19.89 & 20.07 & 20.03 & 18.76 & 20.15 & 19.78 \\ 
            Fixed Query & \textbf{20.13} & 20.03 & 20.14 & 19.88 & 20.16 & 20.25 & 20.16 & 18.74 & 19.84 & 19.93 \\ 
            \rowcolor{mygray}
            \textbf{Learnable Query} & 20.01 & \textbf{20.29} & \textbf{20.39} & \textbf{20.41} & \textbf{20.72} & \textbf{20.81} & \textbf{20.58} & \textbf{19.11} & \textbf{20.53} & \textbf{20.32}\\
            \bottomrule
        \end{tabular}}
    \end{center}
\end{table*}

\begin{figure}[tb]
    \centering
    \begin{minipage}[b]{0.38\textwidth}
        \centering
        \includegraphics[width=\textwidth]{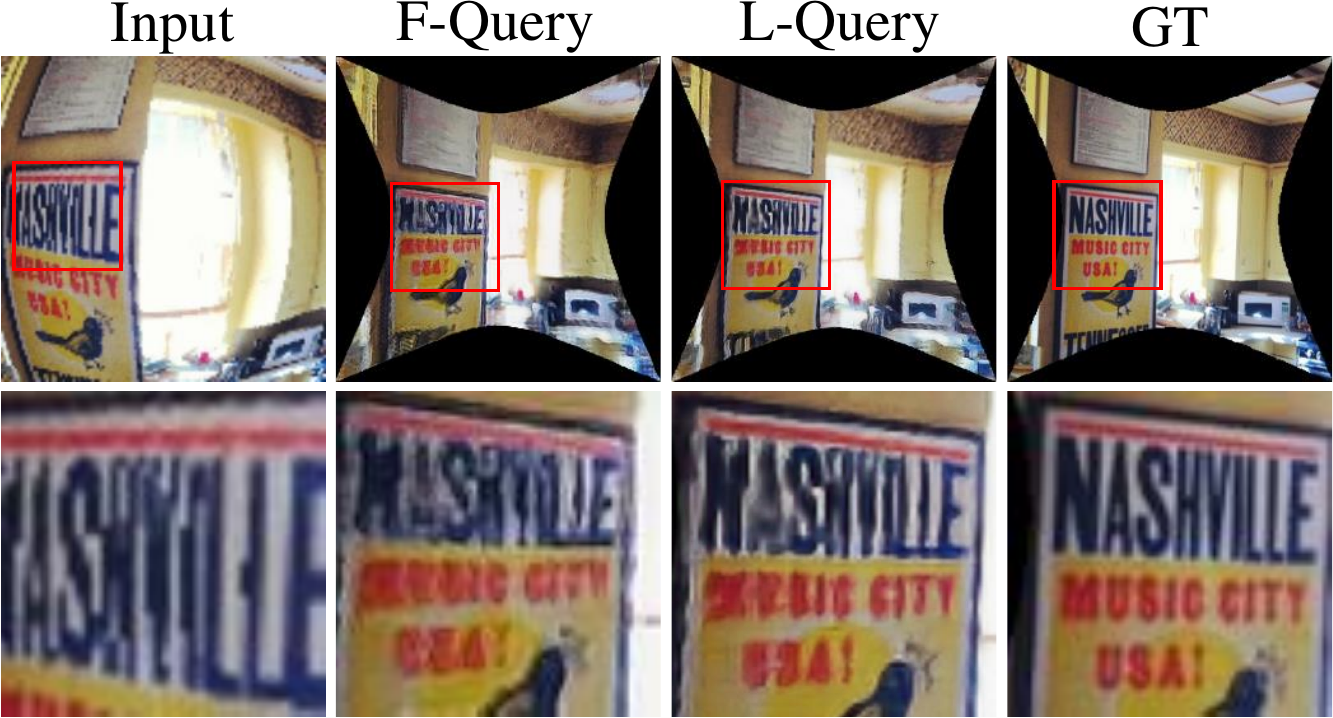}
        \caption{Visual results of different control mechanisms.}
        \label{fig:abl_dlqm}
    \end{minipage}
    \hspace{0.5mm}
    \begin{minipage}[b]{0.57\textwidth}
        \centering
        \includegraphics[width=\textwidth]{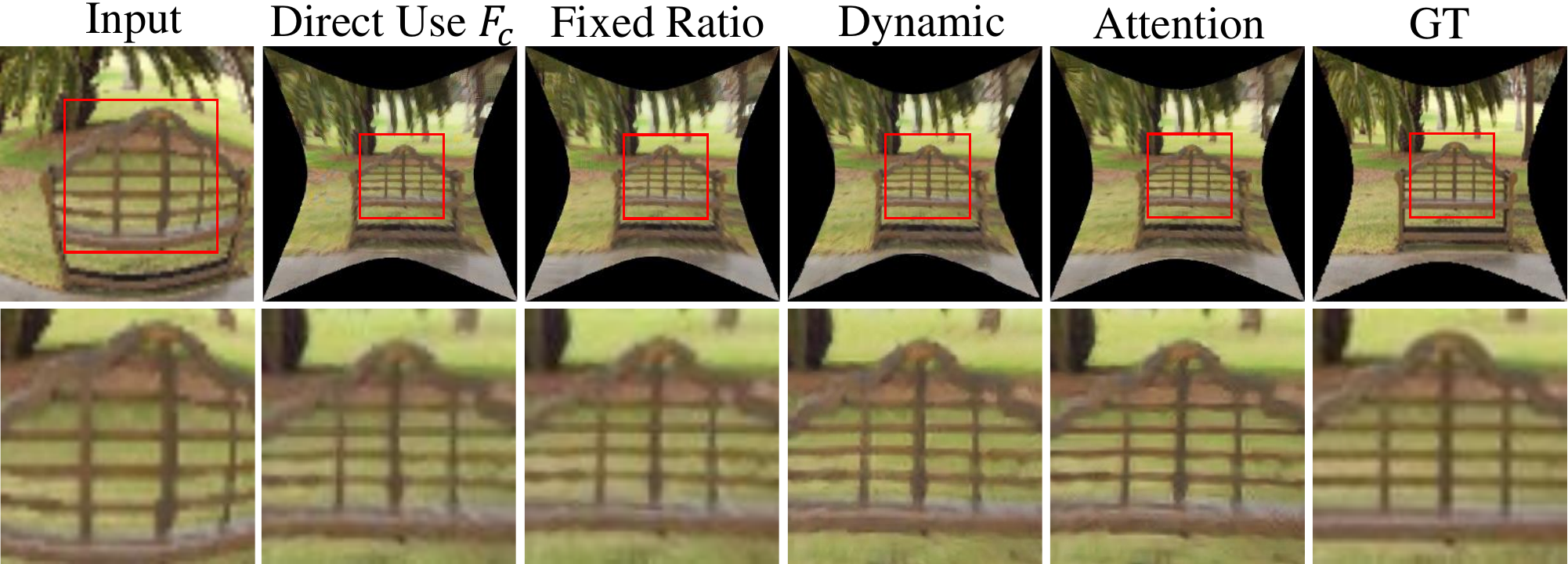}
        \caption{Visual results of different modulation methods.}
        \label{fig:abl_cmb}
    \end{minipage}
\end{figure}

To demonstrate the capability of the CCMB and CAMB in modulating input features with control conditions, we implemented two comparison methods with different modulating approaches: one directly uses the controlled feature (in ~\cref{equ:controlled_fea}), and the other adds controlled feature and original feature in a fixed 1:1 ratio. We constructed different controllable networks using each of these methods.

As shown in~\cref{abl_ccmb} and~\cref{fig:abl_cmb}, retaining original features and fusing them with a 1:1 ratio improved by 0.03 dB, highlighting the importance of preserving original features. The dynamic modulation mechanism boosted performance by 0.06 dB, verifying its effectiveness in dynamic feature modulation. Furthermore, leveraging the control-attention mechanism led to a performance improvement of 0.12 dB, demonstrating the superiority of CAMB for perceiving global distortion distribution, and achieving better uniform global rectification.

\subsubsection{Controllable Rectification Network Architecture.}

\begin{table}[tb]
  \begin{minipage}[tb]{0.46\textwidth}
    \centering
    \caption{Ablation study of different modulation methods. Where Direct Use $F_c$ means directly using the controlled feature $F_c$, Fixed Ratio means adding controlled features and original features in a 1:1 ratio, Dynamic Mechanism means the CCMB and Attention Mechanism means the CAMB.}
    \label{abl_ccmb}
    \begin{tabular}{l c c c c}
        \toprule
        Method & PSNR & SSIM \\
        \midrule
        \rowcolor{mygray}
        Direct Use $F_c$ & 20.14 & 0.655 \\
        Fixed Ratio & 20.17 & 0.658\\
        \rowcolor{mygray}
        Dynamic Mechanism & 20.20 & 0.669\\
        Attention Mechanism & \textbf{20.26} & \textbf{0.671}\\
        \bottomrule
    \end{tabular}
    
  \end{minipage}
  \hfill
  \hspace{0.5mm}
  \begin{minipage}[tb]{0.5\textwidth}
    \centering
    \caption{Ablation study of different network architectures. Where $x\text{C} + (11-x)\text{A}$ denotes that the network uses CCMB in the $x$ layers with larger feature maps and CAMB in the remaining $11-x$ layers with smaller feature maps.}
    \label{abl_net}
    \begin{tabular}{l c c c c}
        \toprule
        Method & Flops(G) & Param(M) & PSNR & SSIM \\
        \midrule
        \rowcolor{mygray}
        PCN~\cite{yang2021progressively} & 12.305 & 35.637 & 17.97 & 0.575 \\
        11C+0A & 12.736 & 37.701 & 20.20 & 0.669\\
        \rowcolor{mygray}
        8C+3A & 13.383 & 46.398 & 20.27& 0.665\\
        6C+5A & 12.353 & 43.244 & \textbf{20.32} & \textbf{0.676} \\
        \rowcolor{mygray}
        4C+7A & 12.538 & 46.994 & 20.31 & 0.670\\
        0C+11A & 15.190 & 51.795 & 20.26 & 0.671\\

        \bottomrule
    \end{tabular}
  \end{minipage}
\end{table}

As described in~\cref{CMB}, rectification networks based on CCMB may ignore the long-range dependencies, while those based on CAMB may ignore the texture details and also increase computational costs. To strike a balance between these two architectures, We incorporate CCMB into the layers with larger feature map size, and integrate CAMB into the layers with smaller feature map size. To validate the effectiveness of this hybrid architecture and find an optimal combination, we conducted performance and computational cost comparisons between pure CCMB, pure CAMB, and multiple hybrid networks, as shown in~\cref{abl_net}. Pure CAMB network (\ie,~Row 6) outperforms pure CCMB network (\ie,~Row 2) in rectification performance but incurs higher computational costs. Hybrid network architectures effectively solve this problem, with higher performance and fewer parameters. After a trade-off between performance and parameters, we empirically choose the $6\text{C}+5\text{A}$ hybrid architecture as our final model. 

\section{Conclusion}

In this paper, we propose the Query-Based Controllable Distortion Rectification network for fisheye images (QueryCDR), which achieves controllable rectification at different distortion degrees without retraining. In particular, we design a series of learnable queries as control conditions to guide the rectification process. Additionally, we design two different controllable modulating blocks, achieving controllable rectification while improving image quality. Extensive experiments have demonstrated the robustness and effectiveness of our QueryCDR. In the future, it is expected to further rectify distortions of different degrees by auto-controlling mechanisms to avoid human involvement. We believe our work provides an effective solution for fisheye camera applications.

\section*{Acknowledgements}
This work was supported in part by the NSFC under Grant 62272376, 62272380 and 62103317, the Key Research and Development Program of Shaanxi Province under Grant 2020ZDLGY04-05, the Fundamental Research Funds for the Central Universities, China (xzy022023051), the Innovative Leading Talents Scholarship of Xi'an Jiaotong University (Corresponding author: Xingsong Hou).

%
%
\bibliographystyle{splncs04}
\bibliography{main}
\end{document}